\documentclass[10pt,twocolumn,letterpaper]{article}
\ifdefined\PtenAnonymous
\usepackage[review]{cvpr}
\else
\usepackage[pagenumbers]{cvpr}
\fi
\usepackage{iftex}
\ifXeTeX
\usepackage{fontspec}
\fi
\usepackage{amsmath,amssymb,booktabs,graphicx,multirow,enumitem,tabularx}
\usepackage{microtype,balance}
\usepackage{caption}
\definecolor{cvprblue}{rgb}{0.21,0.49,0.74}
\ifdefined\PtenAnonymous\PassOptionsToPackage{pagebackref}{hyperref}\fi
\usepackage[breaklinks,colorlinks,allcolors=cvprblue]{hyperref}
\usepackage[capitalize]{cleveref}
\def\paperID{Unassigned}
\def\confName{CVPR}
\def\confYear{2027}

\newcommand{\pp}{\,\mathrm{pp}}
\title{When Better Policies Leave Harder Failures:\\Coverage Shift in VLM Success Monitoring}
\author{Isabel Guan\textsuperscript{1,*}\quad Dingyuan Zhang\textsuperscript{2}\quad Shipeng Lyu\textsuperscript{3}\quad I-Ming Chen\textsuperscript{2}\\
{\small\textsuperscript{1}The Hong Kong University of Science and Technology\quad\textsuperscript{2}Nanyang Technological University}\\
{\small\textsuperscript{3}The Hong Kong Polytechnic University\qquad\textsuperscript{*}Correspondence: \href{mailto:eeguan@ust.hk}{eeguan@ust.hk}}}
\ifdefined\PtenAnonymous
\hypersetup{pdfauthor={}}
\else
\hypersetup{pdfauthor={Isabel Guan, Dingyuan Zhang, Shipeng Lyu, I-Ming Chen}}
\fi
\makeatletter
\appto\@maketitle{\begin{minipage}{\textwidth}
\ifdefined\PtenAnonymous\nolinenumbers\fi
\captionsetup{hypcap=false}
\centering\includegraphics[width=\linewidth]{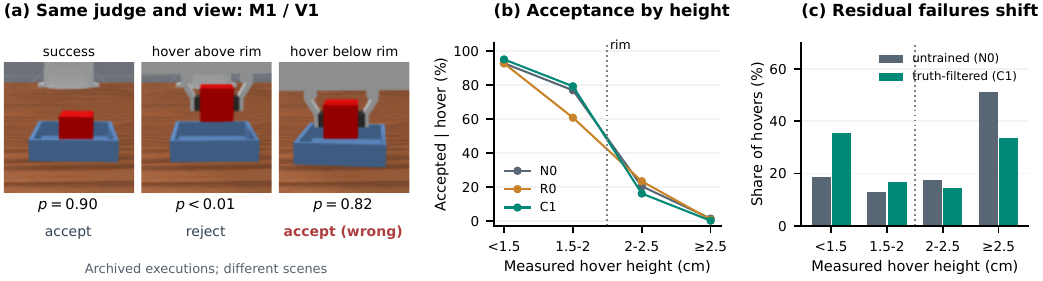}
\captionof{figure}{\textbf{Learning leaves fewer failures, but moves them into a region the judge misses.} (a) Actual archived executions, all judged by M1 at V1; examples are distinct scenes. (b,c) Revised common-reset run: 20 seeds, rounds 1--5, pooled height-bin ratios. Below-rim hovers pass much more often, and truth-filtered residual hovers shift toward this region. Seed-mean success rises from 66.8\% to 97.0\%, conditional hover acceptance from 31.1\% to 50.0\%, while absolute accepted hovers fall from 5.0\% to 0.9\%.}
\label{fig:teaser}
\end{minipage}
\vspace{6pt}
}
\makeatother
\begin{document}
\maketitle
\begin{abstract}
Can a better robot policy become harder to monitor? We study a fixed vision-language model (VLM) that judges success while a policy improves. In a controlled 20-seed tray-placement experiment, training selects only simulator-verified successes. Success rises from 66.8\% to 97.0\%, and accepted hover failures fall from 5.0\% to 0.9\% of executions. Yet the judge accepts 50.0\% of remaining hovers, versus 31.1\% under untrained sampling. These residual failures more often lie below the tray rim, which can conceal the gap beneath a grasped cube. A separate boundary probe accepts 148/160 below-rim hovers and 0/160 above-rim hovers. We call this change in conditional monitoring sensitivity \emph{coverage shift}: learning changes which failures reach a fixed judge. A broader diagnosis covers nine judges and 1,392 stimuli. Judge-filtered experiments across eight settings further show heterogeneous effects of state-dependent acceptance. The results distinguish task improvement from failure detectability and motivate reporting absolute misses, conditional misses, and audit coverage together.
\end{abstract}

\begin{figure*}[t]
\centering\includegraphics[width=\textwidth]{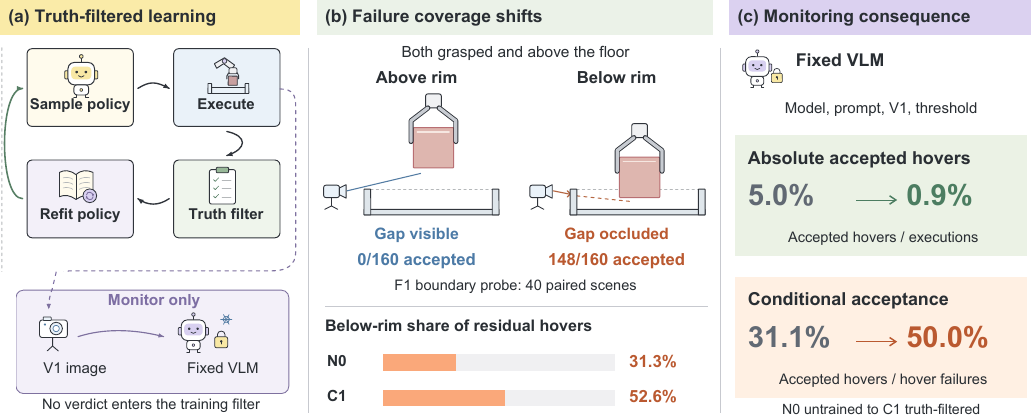}
\caption{\textbf{How successful learning can leave less detectable failures.} (a) C1 fits a controller-parameter distribution using simulator-confirmed successes. VLM verdicts are recorded only for monitoring. (b) The author's schematic shows a grasped cube above and below the tray rim; both remain off the floor and fail the task. F1 counts come from 40 paired scenes. The bars instead use the revised 20-seed audit and report equal-seed means over rounds 1--5. (c) In that audit, accepted hovers decrease per execution but increase as a fraction of hovers. The judge, prompt, camera, and threshold stay fixed. F1 and the revised audit are separate evidence, and the drawing illustrates geometry rather than identifying the judge's internal computation.}
\label{fig:overview}
\end{figure*}
\section{Introduction}
\label{sec:intro}
Robot learning systems increasingly ask vision-language models (VLMs) whether an attempt succeeded. A verdict can select experience for further training, provide a reward, or alert an operator to failure~\cite{du2023success,zhou2024autonomous,lee2026roboreward}. These uses share a requirement: the judge must recognize the failures produced by the policy it serves. Evaluating it on a fixed set of outcomes measures this ability only for the states in that set. As the policy learns, the failures requiring detection may change.

The change can occur while task performance improves sharply. Consider placing a cube in a tray. A cube still grasped above the tray is an obvious failure when its bottom is above the rim. Lowering it below the rim can hide the gap from a low frontal camera, even though it remains grasped and off the floor. If learning removes many high hovers but leaves low hovers, a fixed judge faces fewer failures overall and a more difficult mixture of failures. Both facts matter: the system makes fewer undetected errors per attempt, while its monitor detects a smaller fraction of the errors that remain.

Our principal experiment establishes this distinction without using the judge to train the policy. A truth-filtered arm selects only simulator-confirmed successes. Compared with untrained sampling, it raises success from 66.8\% to 97.0\% and reduces hover incidence from 15.9\% to 1.8\%. Yet the judge's acceptance conditional on a hover rises from 31.1\% to 50.0\%, a paired difference of 18.9 percentage points across 20 seeds. The share of hovers below the rim rises from 31.3\% to 52.6\%. The mechanism overview in \cref{fig:overview} separates the training path, the changed failure geometry, and the two monitoring denominators.

We use \emph{coverage shift} to describe this change in the failure states on which a fixed judge is evaluated. The term identifies a distributional mechanism, not a new learning algorithm. It does not imply that the judge deteriorates on the same images. It also does not require a policy to exploit the judge: simulator truth supplies every training decision in the central comparison. The policy family nevertheless matters. Our controller shares height parameters between released and unreleased actions, and its refitting procedure imposes smoothing and probability constraints. These implementation choices provide a plausible route from successful training to low residual hovers, which our experiments localize but do not fully isolate.

We organize the evidence around three questions. What physical and visual coverage does a static diagnosis provide? How does learning change conditional monitoring sensitivity when truth supplies the labels? When a judge does select training samples, does the assignment of its errors to individual states affect subsequent failures? The last question uses a separate permutation intervention across eight judge--task--view settings. Its results concern the learning signal and cannot be added to the monitoring comparison as an effect decomposition.

The contributions are an empirical counterexample to treating task improvement as preserved monitoring sensitivity, a boundary-based account supported by executed states, and controls that distinguish refitting, truth selection, and state-dependent judge selection. The broader nine-judge diagnosis and eight-setting experiments also expose limits: static prediction and a proposed repair criterion fail their prespecified tests, and several permutation contrasts remain unresolved. Our recommendation is therefore specific: evaluate the failures the policy visits, retain probes near physical success boundaries, and report conditional misses alongside their absolute frequency.

\section{Related Work}
\label{sec:related}
\paragraph{Visual success and reward models.}
SuccessVQA studies task-success detection through visual questions~\cite{du2023success}, and Luo examines sensitivity to visual distribution shift~\cite{luo2024success}. VLM-RMs and RoboCLIP derive rewards from pretrained vision--language representations~\cite{rocamonde2024vlm,sontakke2023roboclip}. RoboReward broadens robotic reward modeling with negative examples~\cite{lee2026roboreward}. REFLECT uses multisensory summaries to explain failures~\cite{liu2023reflect}; AHA trains failure detection and explanation~\cite{duan2025aha}; FailBench studies natural manipulation failures and visual evidence~\cite{navasardyan2026failbench}. Our diagnosis holds an executed physical state fixed across views, then asks how learning changes the states that a monitor must judge.

\paragraph{Learning from selected experience.}
SOAR autonomously collects and evaluates robot experience~\cite{zhou2024autonomous}, while RoboCat trains on additional self-generated data~\cite{bousmalis2023robocat}. We isolate the selection step through a small controller distribution. Selection followed by distribution fitting is classical in the cross-entropy method~\cite{deboer2005tutorial}. Erroneous pseudo-labels can reinforce errors~\cite{arazo2020pseudo}; our truth-filtered comparison instead supplies correct labels and measures changes in a separate monitor. This also relates to performative prediction~\cite{perdomo2020performative}: decisions can change an evaluated distribution, including when the evaluator does not direct those decisions.

\paragraph{Imperfect evaluation under optimization.}
Goodhart-style failures and reward overoptimization concern a proxy's relationship to its target under selection~\cite{manheim2018goodhart,gao2023scaling,wen2025rethinking,kim2025rethinking}. Imperfect verifiers limit resampling~\cite{stroebl2024limits}, and verifier error structure affects arithmetic reinforcement learning and language-model cascades~\cite{egashira2026delay,rajput2026cheap}. CriticHack is the closest robotic comparison: it studies visual reward optimization, diffusion-policy fine-tuning, an exponential-tilt account, and a frozen verifier intervention~\cite{luo2026critichack}. Its results include rising wrong-object failures alongside rising success. Our central observation instead combines decreasing absolute accepted failures with increasing conditional acceptance under truth-based training. The subsequent permutation study concerns threshold filtering and parametric refitting. CriticHack was available during the later interpretation and audit; the supplement preserves chronology and evidence scope without claiming independent conceptual priority.

\section{Evaluation Framework}
\label{sec:testbed}
\subsection{Task performance and monitoring sensitivity}
For a policy-induced distribution of physical outcomes $d_t(x)$, let $S$ denote true success and $B$ a specified failure set. A fixed judge returns acceptance $a_j(x)$ under a fixed camera, prompt, and threshold. We measure
\begin{align}
 s_t&=\Pr_{d_t}(S), & b_t&=\Pr_{d_t}(B),\nonumber\\
 m_t&=\mathbb E_{d_t}[\mathbf1_B a_j(x)], & c_t&=m_t/b_t\quad(b_t>0).
 \label{eq:metrics}
\end{align}
Here $s_t$ is task success, $b_t$ is failure incidence, $m_t$ is accepted failures per execution, and $c_t$ is acceptance conditional on the specified failure. Conditional acceptance is a miss rate for that failure set. The identity $m_t=b_t c_t$ explains how a falling failure incidence can outweigh a rising conditional miss rate. It holds within a distribution or seed; multiplying separately averaged seed ratios need not recover the mean product.

The conditional rate can change even if the judge is fixed:
\begin{equation}
 c_t=\int_B a_j(x)\,d_t(x\mid B).
 \label{eq:shift}
\end{equation}
Learning changes the mixture of states inside $B$. This distinction matters even within a named failure type such as hover, because height and occlusion can vary within that type. A static error map estimates responses on its sampled support. Transfer to a later policy depends on how well that support represents the policy's residual failures. This is the coverage question tested below.

\subsection{Physical tasks and policy updates}
We use a Franka Panda arm and 4\,cm cubes in ManiSkill3~\cite{tao2024maniskill3}. The robot stacks a red cube on a green cube (\emph{stack}) or places it inside a tray (\emph{tray}). The tray has a 9\,cm inner side, a 5\,mm floor, and a rim 20\,mm above the floor. Every evaluated outcome follows controller execution. Success requires the simulator's placement predicate, a static cube, and no grasp. The judge never defines physical truth.

The policy $\pi_t(\theta)$ samples parameters of a scripted controller: a table-versus-goal branch, horizontal offsets, goal-release height, release, and retreat. The fitted family uses Bernoulli branches, Gaussian offsets, and a log-normal height clipped to $[2,50]$\,mm. Each round executes 100 parameter samples in fresh scenes, selects them with a binary filter $a_i$, and refits:
\begin{equation}
 \pi_{t+1}=\operatorname{Fit}_{\mathcal P}\{(\theta_i,a_i)\}_{i=1}^{100},
 \quad\text{targeting }\max_{\pi\in\mathcal P}\sum_i a_i\log\pi(\theta_i).
 \label{eq:refit}
\end{equation}
The implementation adds Bernoulli smoothing, clips probabilities to $[.02,.98]$, and constrains variances. With no accepted samples the policy stays unchanged. The initial natural distribution lies outside the fitted family. Consequently, even fitting all samples can change the distribution. The supplement specifies the update; \cref{eq:refit} is its objective, not an exact unconstrained density transformation.

\paragraph{Controls.}
N0 samples the initial distribution without learning. R0 refits all samples. C1 keeps simulator-confirmed successes. C3 keeps judge-accepted samples. C2* permutes judge acceptance within the true-success and true-failure strata of its own current pool, preserving that pool's TP, FP, and acceptance count while changing which states receive acceptance. A calibrated-noise control C2 instead uses fixed N1 TPR/FPR. Safeguards tighten the score threshold (C6), require an additional checking view (CV), or veto a positive grasp probe (C4). C3t keeps the highest-scoring samples while matching a safeguard's retained count. Truth-dependent arms are experimental controls.

\subsection{Experiments and inferential units}
The historical loop crosses two judges, two tasks, and two deployment views, producing eight settings with five policy seeds each. It fixes target failure sets and safeguards from the diagnostic map. Six rounds ($t=0,\ldots,5$) precede 100 fresh truth-scored final executions. The later C3/C2* experiment uses the same budget and independent final evaluation. Paired seeds share scenes and sampling streams, but their states can diverge after filters differ. C2* matches confusion counts within its own pool, not across complete C3/C2* trajectories.

The central monitoring audit jointly runs N0/R0/C1 in M1/tray/V1 with a common reset and 20 seeds (5--24). Each seed's endpoint pools rounds 1--5 before forming ratios. This is a measurement during training, not an independent final test. It followed earlier observations, reused F1 seed numbers, and is not an independent confirmatory replication. We report equal-seed means and paired two-sided 95\% $t$ intervals. All 20 seeds meet the minimum of three pooled hovers per arm; C1 retains 4--15 per seed.

The original H3, T1--T4, and boundary follow-up F1 had local dated or hash-locked protocols before their respective data, rather than an external registry. Their original decisions and deviations remain visible. The revised mechanism and permutation experiments contribute 92,000 formal arm-round records: 36,000 and 56,000 respectively. Seeds, rather than these dependent rows, are the inferential units. C2* intervals use 10,000 paired-bootstrap resamples and are descriptive, without multiplicity adjustment. Pilots and unfinished experiments provide no formal evidence for the claims here.

\section{Static Diagnosis and Boundary Coverage}
\label{sec:diagnosis}
\subsection{Holding truth fixed while changing the view}
The diagnostic set contains 12 scenes per task, with 14 stack and 15 tray design conditions per scene: 348 executed end states. Rendering each at $512\times512$ from V1 (front-low), V2 (oblique), V3 (top-down), and V4 (level side) gives 1,392 images. Seeds 0--7 form the main split, 8--11 the held-out split, and 20--31 a second diagnostic set. Natural pools N1 (100 states per task) and N2 (300 per task) come from the initial controller distribution. N1 calibrates controls; N2 tests accepted-failure composition. Views share physical truth, whereas the natural pools change the sampled states.

Ordered simulator predicates distinguish \emph{hover} (grasped, cube bottom more than 6\,mm above support), \emph{held\_on} (grasped at or below that threshold), \emph{behind} (ungrasped and flat on the table behind the goal), \emph{edge} (unsuccessful near-goal placement not flat outside it), and \emph{beside} (other ungrasped failures). Exact geometry and precedence appear in \cref{sec:supp_stimuli}. Error-map and composition analyses exclude held\_on. Physical success, noise calibration, and the historical T4 denominator follow their disclosed frozen definitions.

Nine judges participate. J1 is Qwen3-VL-8B-Instruct~\cite{bai2025qwen3vl}; M1 is Qwen3.8-27B~\cite{qwen2026qwen38}; J2 is RoboReward-8B; J4 is a logistic head on frozen DINOv2 features~\cite{oquab2023dinov2}. Five closed models appear in \cref{tab:cues}. These names identify archived experiments, not current endpoint rankings. J4 uses V2, Opus uses V1/V2, and the others use all four views. J2 receives 16-frame clips and a calibrated progress threshold. Other VLMs receive one final frame and a frozen yes/no question. J1/M1 threshold normalized Yes/No probabilities at $.5$. Closed answers map to Yes ($1$), No ($0$), or unparsed ($.5$); the main rule accepts the latter. Treating abstentions as rejections changes no H3 verdict. Four separate cue questions probe reported visual evidence.
\begin{figure*}[t]
\centering\includegraphics[width=\textwidth]{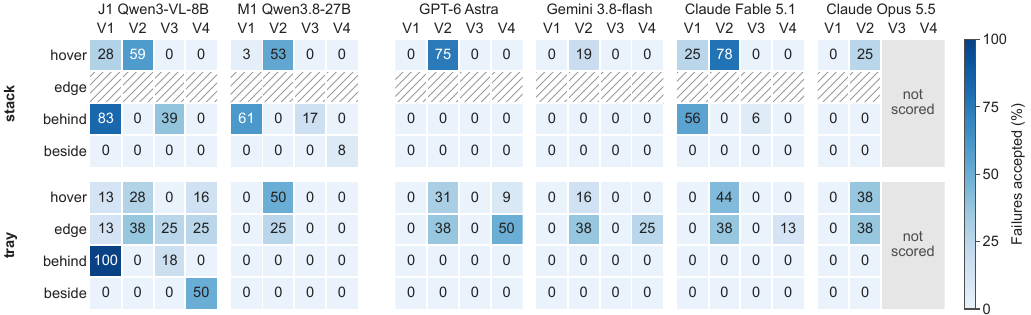}
\caption{\textbf{Errors depend on failure type and view.} Main-split acceptance percentages for six selected judges, reproduced from the archived diagnostic figure. Stack is above tray. Held\_on is excluded; the missing stack-edge type is hatched and uses the pooled task failure rate in prediction. Claude Opus 5.5 has two views only. Shared physical states across views make camera-dependent ambiguity visible. The full nine-judge protocol and coverage are in the supplement.}
\label{fig:errormaps}
\end{figure*}

\paragraph{Error maps and their limits.}
For states $X_f$ of failure type $f$ in the main split,
\begin{equation}
 A_j(f,v)=|X_f|^{-1}\sum_{x\in X_f}\mathbf1[p_j(x,v)\ge\gamma_j].
 \label{eq:errormap}
\end{equation}
Types with fewer than four states use pooled failure acceptance at that view. Given target-pool type counts $n(f)$, the predicted accepted-failure composition is
\begin{equation}
 \widehat q_{j,v}(f)\propto n(f)A_j(f,v).
 \label{eq:h3pred}
\end{equation}
B1 uses $n(f)$ alone, and B2 uses $n(f)A_j(f,\mathrm{V2})$. The H3 criterion requires Spearman correlation at least $.5$ and mean KL below both baselines, with $10^{-3}$ additive smoothing. This predictor uses target-pool type counts; it does not forecast a future policy solely from its initialization.

Errors vary strongly with type and view (\cref{fig:errormaps}). Hover acceptance concentrates at V2 and behind acceptance at V1. M1 has lower aggregate false acceptance than J1 on both diagnostic sets (9.8/10.8\% versus 21.1/21.7\%), yet V2 hovers account for 62\% of M1's accepted N2 failures versus 28\% for J1. This is a composition comparison, not a scaling law. Cue regressions reveal that some judges report a grasp yet still accept, but do not identify internal reasoning. H3 passes its point-estimate rule for five of eight testable judges. Later scene-bootstrap intervals provide no judge a positive KL-advantage lower bound against both baselines. J4 is excluded because its B2 and map are identical.
\begin{table}[t]
\centering\footnotesize\setlength{\tabcolsep}{3pt}
\begin{tabular}{@{}lcc@{}}\toprule
Judge & Stack grasp weight & Tray grasp weight\\\midrule
Gemini 3.6-flash & $.10\ [.02,.20]$ & $-.29\ [-.47,-.17]$\\
GPT-6 Astra & $-.02\ [-.22,.17]$ & $-.05\ [-.24,.12]$\\
Claude Fable 5.1 & $.09\ [.02,.15]$ & $.02\ [-.11,.32]$\\
Claude Opus 5.5 & $-.32\ [-.49,-.21]$ & $-1.47\ [-2.33,-1.06]$\\
Gemini 3.8-flash & $-.41\ [-.56,-.32]$ & $-1.41\ [-2.20,-1.26]$\\\bottomrule
\end{tabular}
\caption{Descriptive logistic cue weights, with 95\% scene-bootstrap intervals (1,000 resamples). Ideal-observer weights are $-.86/-.57$ for stack/tray; for Opus's two-view subset they are $-.78/-.53$. The registered weak-use criterion is an upper bound above $-.3$ with a negative ideal weight, not a test of perceptual or causal equivalence.}
\label{tab:cues}
\end{table}

\subsection{A missing region near the tray rim}
\label{sec:res_t1}
The registered T1 transfer test fails in the historical loop. Across eight settings, predicted and observed composition correlate at $.821$, but map KL is $.383$, worse than B1's $.342$ and better only than B2's $.792$. At V2, KL is $.021$--$.044$ and B2 equals the map. At V1, the map underestimates hovers in every setting. M1/tray/V1 is especially revealing: no diagnostic failure is accepted, yet hovers make up 81\% of accepted loop failures. Its zero forecast becomes uniform under frozen smoothing. Dropping that setting or substituting B1 reverses the aggregate comparison, but those post hoc sensitivities do not change T1's failed verdict.
\begin{figure*}[t]
\centering\includegraphics[width=.96\textwidth]{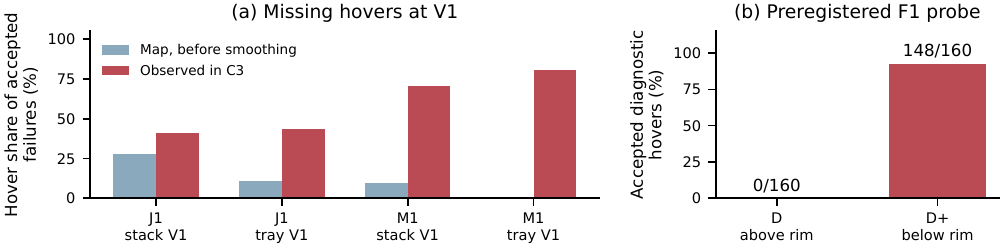}
\caption{\textbf{A coverage gap near the success boundary.} Left: historical T1 hover composition at V1; map values precede smoothing (the all-zero M1/tray/V1 prediction is uniform after smoothing for KL). Right: the preregistered F1 diagnostic probe accepts 0/160 above-rim hovers (D) and 148/160 below-rim hovers (D+), across 40 valid scenes. This is a state-coverage comparison, not evidence that the judge changed over time.}
\label{fig:coverage}
\end{figure*}

The missed coverage is geometric. Original diagnostic hovers in this setting lie above the rim, whereas the loop also visits lower hovers. After exploratory inspection, the registered F1 probe executes above-rim D and below-rim D+ cases in the same 40 valid scenes. The fixed judge accepts $0/160$ D hovers and $148/160$ D+ hovers; the one-sided 95\% scene-level lower bound on their difference is $.891$. Historical C3 hover acceptance over rounds 1--5 is $.480$, with one-sided 97.5\% lower bound $.435$, above the conservative D scene bound $.088$. These measurements identify a gap in sampled support. They neither show that the judge changed over time nor measure the later common-reset audit, which is reported separately.

The original map sets each loop cell's target failure set and safeguards. Success-versus-dominant-failure AUROC defines separable ($\ge .95$), inseparable ($<.85$), and intermediate cells. If every eligible type has acceptance below $.10$, the cell is labelled no-blind-spot. M1/tray/V1 receives that label. Its later hover endpoint is a frozen control type; the boundary finding does not retrospectively convert it into a diagnosed blind spot.

\section{Learning Changes the Failures to Be Judged}
\label{sec:results}
\subsection{Truth filtering improves behavior but raises conditional misses}
\label{sec:res_mechanism}
\begin{figure}[t]
\centering\includegraphics[width=\linewidth]{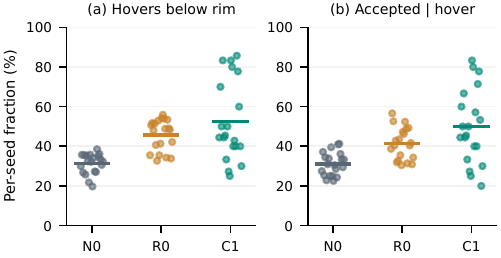}
\caption{\textbf{The shift appears across seeds.} Each point is one seed's ratio pooled over rounds 1--5; horizontal lines are equal-seed means. Truth filtering (C1) leaves 4--15 hovers per seed. The untrained (N0), refit-only (R0), and truth-filtered arms use the revised common-reset protocol.}
\label{fig:mechanism}
\end{figure}

The common-reset comparison tests the central monitoring question. In M1/tray/V1, C1 uses only simulator truth for selection, while the fixed judge records verdicts. Over rounds 1--5, success rises from N0's 66.8\% to 97.0\%, and hover incidence falls from 15.9\% to 1.8\%. Accepted hovers per execution consequently fall from 5.0\% to 0.9\%. The monitor nevertheless accepts a larger share of hovers: 50.0\% under C1 versus 31.1\% under N0. The paired increase is $18.9\pp$ with two-sided 95\% interval $[9.7,28.1]$ (\cref{tab:mechanism,fig:mechanism}). Better physical performance and lower conditional sensitivity coexist.
\begin{table}[t]
\centering\footnotesize\setlength{\tabcolsep}{3pt}
\begin{tabular}{@{}lrrr@{}}\toprule
Metric (\%) & N0 & R0 & C1\\\midrule
True success & 66.81 & 66.69 & 96.99\\
Hover incidence & 15.92 & 17.64 & 1.75\\
Accepted given hover & 31.09 & 41.39 & 49.98\\
Below rim given hover & 31.26 & 45.72 & 52.65\\
Hover and accepted / execution & 4.97 & 7.21 & 0.86\\\midrule
Hover count & 1,592 & 1,764 & 175\\
Accepted hover count & 497 & 721 & 86\\\bottomrule
\end{tabular}
\caption{Common-reset mechanism run, M1/tray/V1. Percentages are equally weighted seed-rate means (20 seeds, 500 observations per seed and arm, rounds 1--5). Counts pool seeds. In particular, $86/175=49.14\%$ differs from the seed mean $49.98\%$.}
\label{tab:mechanism}
\end{table}

\paragraph{Where the remaining failures occur.}
The equal-seed below-rim share rises from 31.3\% to 52.6\%, a paired increase of $21.4\pp$ $[11.7,31.1]$. Below-rim hovers are accepted much more frequently than higher hovers. Descriptive pooled counts are $431/499$, $640/795$, and $82/91$ below the rim for N0, R0, and C1; at or above it they are $66/1093$, $81/969$, and $4/84$. The height distributions in \cref{fig:teaser} and the separate F1 probe support a coverage account: the residual distribution puts more mass in a region where the fixed view hides useful evidence. They do not establish equal response functions within coarse height strata or uniquely identify the judge's visual computation.

\paragraph{What refitting contributes.}
R0 fits every sample and uses neither truth nor judge selection. It raises conditional hover acceptance by $10.3\pp$ $[5.2,15.4]$ relative to N0. C1 adds $8.6\pp$ $[1.4,15.8]$ relative to R0. For the below-rim share, C1 minus R0 is $6.9\pp$ $[-0.7,14.6]$, leaving that contrast unresolved at the two-sided level. Refitting alone therefore changes the monitored distribution, while truth selection further changes conditional acceptance. These contrasts do not isolate each numerical ingredient of the update.

The initial tray command height is uniform on $[4,50]$\,mm; the fitted height family is clipped log-normal. R0 combines family projection with finite samples, recursive estimation, smoothing, and constraints. In every post-initial C1 seed-round, release probability reaches the $.98$ cap. Its remaining non-release branch shares height parameters with successful releases. Training toward lower successful placements can thus also lower residual hovers, although no hover enters C1's selected data. This coupling is a plausible implementation mechanism. Removing the cap, separating branch parameters, or matching the initial family would be needed to isolate it experimentally. Historical F1 discrepancies and the revised reset checks remain documented separately.

\begin{table*}[t]
\centering\small\setlength{\tabcolsep}{7pt}
\begin{tabular}{@{}llrrrrr@{}}\toprule
Cell & Frozen failure set & C3 failure & C2* failure & C3$-$C2* [95\% interval] & C3 success & C2* success\\\midrule
M1 / stack / V1 & behind & 0.4 & 0.2 & $+0.2\ [+0.0,+0.6]$ & 95.2 & 95.2 \\
M1 / stack / V2 & hover & 5.6 & 2.0 & $+3.6\ [+1.6,+5.6]$ & 87.0 & 93.6 \\
M1 / tray / V1 & hover & 3.2 & 3.0 & $+0.2\ [-3.2,+3.2]$ & 94.4 & 95.0 \\
M1 / tray / V2 & edge, hover & 16.0 & 10.0 & $+6.0\ [+4.6,+7.8]$ & 83.4 & 88.2 \\
J1 / stack / V1 & behind, hover & 4.2 & 2.6 & $+1.6\ [-1.4,+5.4]$ & 92.4 & 94.4 \\
J1 / stack / V2 & hover & 5.6 & 2.4 & $+3.2\ [+0.8,+5.8]$ & 85.6 & 93.8 \\
J1 / tray / V1 & behind & 15.4 & 1.2 & $+14.2\ [+6.0,+21.4]$ & 75.8 & 92.8 \\
J1 / tray / V2 & edge, hover & 3.4 & 2.8 & $+0.6\ [-0.2,+1.4]$ & 96.4 & 97.0 \\

\bottomrule\end{tabular}
\caption{Revised C3/C2* experiment: independent final evaluation, five paired seeds per cell, 100 rollouts per arm and seed. All values are percentages except differences in pp. Intervals are descriptive paired-bootstrap intervals, not multiplicity-adjusted. M1/tray/V1's hover endpoint is a control type; no blind spot was detected there in the original main-set map.}
\label{tab:c2star}
\end{table*}
\begin{figure*}[t]
\centering\includegraphics[width=\textwidth]{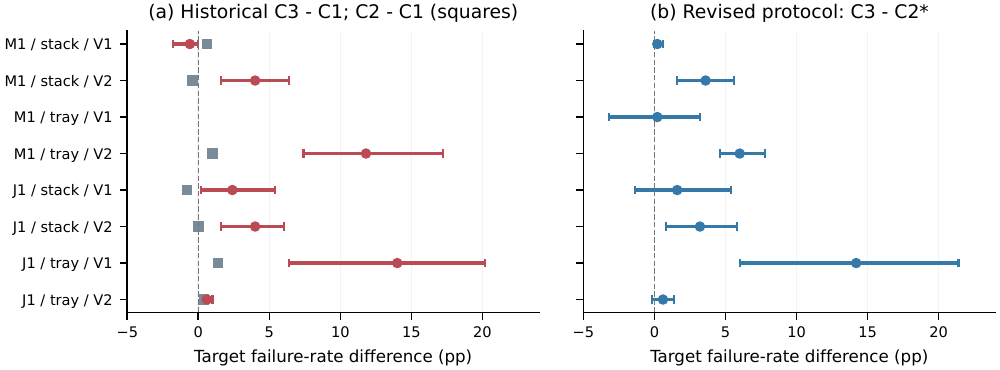}
\caption{\textbf{Two controls for error structure, with different protocols.} (a) Historical final target failure differences: C3 minus C1 (circles and seed-bootstrap intervals) and C2 minus C1 (squares, means only); the no-blind-spot cell has no T2 test. (b) Later C3 minus C2* differences. Each interval uses five paired seeds and is unadjusted. Panels must not be added or interpreted as a single decomposition.}
\label{fig:structure}
\end{figure*}

\subsection{When the judge selects data, state assignment matters}
\label{sec:res_structure}
The monitoring result does not require erroneous training labels. A separate question asks what those labels do when used for selection. Historical T2 compares final target failures under C3 and C1. Six of seven diagnosed blind-spot cells satisfy its rule, exceeding the required five, with C3--C1 differences of $0.6$--$14.0\pp$. M1/stack/V1 fails. Fixed calibrated noise C2 differs from C1 by $-0.8$ to $+1.4\pp$. In six of seven cells, C3 still reduces target failures from initialization by $2.2$--$14.6\pp$; only J1/tray/V1 increases them, by $7.8\pp$. The common effect is poorer removal relative to truth filtering, rather than an absolute rise in failures.

The later C2* intervention breaks the assignment of acceptance to individual states within truth strata. Four of eight descriptive C3--C2* intervals are strictly positive, three cross zero, and one touches zero (\cref{tab:c2star,fig:structure}). In J1/tray/V1, behind failures are 15.4\% under C3 versus 1.2\% under C2*, while success is 75.8\% versus 92.8\%. M1/tray/V1 has little resolved difference. These results support heterogeneous effects of state assignment. Because each trajectory generates its own pool, later error rates, retained counts, and learning progress may also differ. C2* preserves its own pool's confusion counts, not a trajectory-wide error rate shared with C3.

Historical T4 asks a narrower cross-model question. Designated blind-spot types comprise $.570$ of M1's accepted failures and $.511$ of J1's, a difference $.059$ $[-.020,.126]$ whose lower bound exceeds the $-.05$ non-inferiority margin. Excluding held\_on changes the shares to $.915$ and $.705$. This result concerns failure composition. Differences in model generation, size, and denominators prevent a general inference about scaling or absolute reliability.

\subsection{Additional evidence can help; filtering more strictly can harm}
\label{sec:res_repairs}
The historical checking-view control reaches 93.8--97.4\% final success across eight cells, within $3.2\pp$ of C1's 97.0\% (\cref{tab:repairs}). J1/tray/V1 improves from C3's 75.6\% to CV's 94.0\%. CV obtains another view and makes a second query after first-view acceptance. The result therefore supports additional visual evidence under that budget, without establishing equal-cost superiority.
\begin{table}[t]
\centering\footnotesize\setlength{\tabcolsep}{3pt}
\begin{tabular}{@{}lrrrrr@{}}\toprule
Cell & C3 & C4 & C6 & C3t & CV\\\midrule
J1 / stack / V2 & 85.6 & 96.8 & 97.0 & 97.0 & 97.0 \\
M1 / stack / V2 & 87.0 & 97.8 & 97.6 & 97.6 & 97.4 \\
M1 / tray / V2 & 84.6 & 91.6 & 95.4 & 95.4 & 95.6 \\
J1 / tray / V1 & 75.6 & 84.4 & 89.2 & 81.2 & 94.0 \\
J1 / tray / V2 & 96.4 & 96.4 & 97.8 & 96.6 & 97.4 \\
M1 / stack / V1 & 95.2 & 97.2 & 97.8 & 95.6 & 94.8 \\
J1 / stack / V1 & 91.4 & 71.0 & 72.8 & 72.8 & 93.8 \\
M1 / tray / V1 & 95.6 & 97.2 & 94.6 & 94.6 & 97.4 \\

\bottomrule\end{tabular}
\caption{Historical final true success (\%, five-seed means). C1 is 97.0\% throughout; C2 ranges from 94.6--96.4\%. C3t equals C6 by construction except in the three inseparable cells. All filters use the same training budget; CV has extra visual information and query cost.}
\label{tab:repairs}
\end{table}

The prespecified T3 repair rule still fails overall. In three separable cells, stricter threshold C6 improves success by 10.6--11.4\,pp, but C3t selects the same highest-scoring samples, making non-inferiority to that control uninformative. None of the three inseparable cells meets T3. In J1/tray/V1, CV minus C6 is $4.8\pp$ $[-0.6,10.2]$; the other two cells show no CV improvement over C6. In intermediate J1/stack/V1, a $.995$ threshold retains about one-third of diagnostic successes and lowers final success from 91.4\% to 72.8\%. The grasp veto lowers it to 71.0\%. A stricter acceptance rule can discard useful successes and impair learning.

\paragraph{Why a static weighting model is insufficient.}
A later exploratory target-failure/success acceptance ratio correlates with C3--C2* effects at $.407$ across eight cells, compared with $.623$ for static FPR. Repeated weighting by fixed type acceptance yields mean trajectory error $5.1\pp$, versus $3.9\pp$ using round-0 on-policy acceptance and $7.8\pp$ for no change. It predicts 0.3\% final target failures for M1/tray/V2, where 16.0\% occur. These post hoc comparisons are not an independent method contest. Exact one-step filtering identities do not justify holding within-type acceptance fixed across repeated constrained refits. The supplement gives the derivation and negative results.

\balance
\section{Discussion and Limitations}
\label{sec:discussion}
The practical implication is to assess what an improved policy still gets wrong. In the central experiment, the policy succeeds more often and produces fewer undetected hovers per execution. Conditional monitoring nonetheless weakens because the remaining hovers occupy a different part of state space. Reporting either denominator alone would omit part of this result. A useful audit should give task success, failure incidence, absolute accepted failures, and conditional acceptance for specified failure sets, together with counts and the states sampled.

\paragraph{Coverage is a measurement problem.}
A type name such as hover does not guarantee visual similarity. Boundary probes make height and occlusion explicit, while truth-fixed views reveal whether additional evidence changes a verdict for the same outcome. Our F1 probe diagnoses a missing region; it does not certify coverage of all later failures. Likewise, the checking view helps under the tested query and camera budget but does not establish an optimal auditing strategy. Efficient choices of states, labels, and additional views remain open.

\paragraph{Scope and alternative explanations.}
The study uses two simulated tasks, scripted manipulation, and a low-dimensional policy distribution. The central truth-filtered effect is established in one judge--task--view setting using a revised protocol developed after earlier observations. Seed reuse and imperfect historical reproduction limit confirmation. Refitting, family mismatch, shared parameters, clipping, and finite samples remain intertwined. If removing the release cap or separating release and hold heights removes the effect, the explanation should narrow accordingly. A neural policy could lack this coupling; the one-seed diffusion pilot does not validate transfer.

Only J1/M1 participate in the completed multi-seed loops. Closed judges contribute static evidence with unequal input coverage and provider/version limitations. C2* has five paired seeds per cell and unadjusted intervals across related settings. Simulator truth and freely rendered views simplify supervision relative to deployment. Independent prospective prediction, neural-policy confirmation, real-image boundary tests, and equal-label-budget auditing have no completed formal evidence here. The failed static and repair criteria constrain the conclusions alongside the positive results.

\section{Conclusion}
An improving policy can leave fewer failures that a fixed visual judge finds harder to detect. Truth-filtered training in our controlled tray task reduces absolute accepted hovers while increasing acceptance conditional on a hover. The residual failures concentrate below the tray rim, where the fixed view provides weaker evidence of the physical failure. Separate permutation experiments show that state-dependent judge acceptance can also affect learning, with substantial variation across settings. Monitoring quality should therefore be evaluated on the policy's residual failures and reported together with their absolute frequency.

\label{maintextend}
\clearpage
\nobalance
{\small\bibliographystyle{ieeenat_fullname}\bibliography{main}}
\clearpage
\appendix
\begin{center}\Large\bfseries Supplementary Material\end{center}
\noindent This supplement documents the completed evidence and its limitations. It separates the historical main experiment and F1 from the later common-reset mechanism and C2* protocols.

\begin{figure*}[t]
\centering\includegraphics[width=\textwidth]{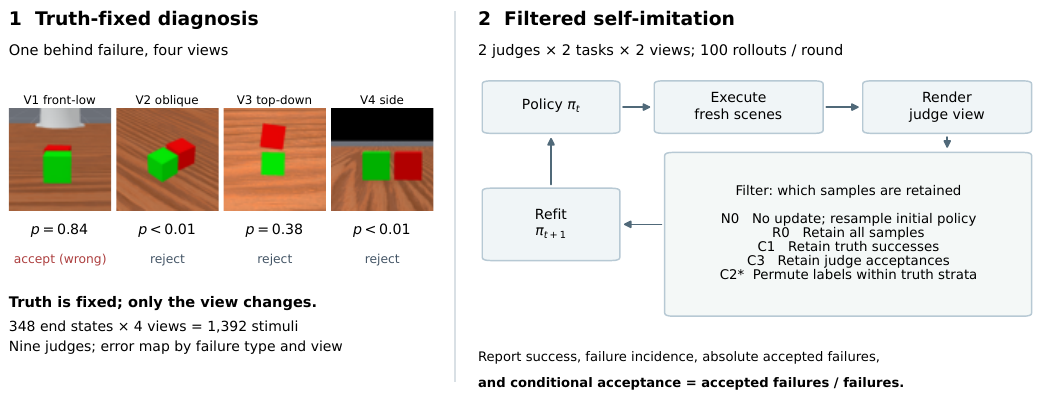}
\caption{\textbf{Multi-view diagnosis and the complete training controls.} Left: one executed stacking failure across four views with archived M1 scores. Right: the filtering loop and controls used across the eight settings. This broader overview complements the C1 mechanism schematic in \cref{fig:overview}. Six training rounds precede independent final evaluation in the historical and permutation loops; the common-reset mechanism audit instead pools rounds 1--5.}
\label{fig:supp_overview}
\end{figure*}

\section{Stimulus design and failure predicates}
\label{sec:supp_stimuli}
This section specifies the stimuli of \cref{sec:diagnosis} exactly.
Lengths are in metres unless stated otherwise and angles in radians, in the simulator's world frame: $+x$ points away from the robot, the table top is at $z=0$, and a cube resting on the table has its centre at $z=0.02$.
$\mathcal{U}(a,b)$ is a uniform draw; every draw comes from NumPy's default generator with the seed given.

\paragraph{Scenes.}
The goal is the green cube (stack) or the open tray (tray), a single kinematic body made of a 5\,mm floor and four 5\,mm thick walls that rise 2.5\,cm above the table around a $9\times9$\,cm interior.
A scene seed $s$ fixes, through a generator seeded with $s$:
the red cube's start position, $x\sim\mathcal{U}(-0.12,-0.04)$ and $y\sim\mathcal{U}(-0.18,0.18)$, with $y$ redrawn as $\operatorname{sign}(y)\,\mathcal{U}(0.08,0.18)$ whenever $|y|<0.08$, so that the cube starts on the robot's side of the goal with $|y|\ge0.08$;
the red cube's yaw, $\mathcal{U}(-0.17,0.17)$;
the goal centre, $(0.05,0)$ plus a jitter of $\mathcal{U}(-0.01,0.01)$ per axis;
and the goal yaw, $\mathcal{U}(-0.17,0.17)$.
The direction $\phi\sim\mathcal{U}(0,2\pi)$ of the release offsets in the design conditions comes from a generator seeded with $10{,}000+s$, and the lighting preset is $s \bmod 2$ (\cref{tab:views}).
The arm starts from a fixed joint configuration without noise.
Physics runs on the CPU, and the end-effector controller always uses ManiSkill's pure-PyTorch kinematics path.

\paragraph{Controller.}
The controller approaches each waypoint in end-effector position steps of at most 1.5\,cm per axis until the gripper centre is within 3\,mm (at most 120 steps per waypoint).
It moves to 10\,cm above the red cube, descends to 3\,mm below the cube's centre, closes the gripper for 12 steps, lifts to $z=0.15$, moves above the release point $g+(\Delta x,\Delta y)$, where $g$ is the actual goal centre and the offset is taken along the world axes, and descends to $z_{\mathrm{s}}+0.02+h$, where $z_{\mathrm{s}}$ is the top of the support (the green cube or the tray floor, or the table for placements on the table), so that the cube bottom is nominally $h$ above it.
After 4 more steps it either releases or keeps holding.
A release opens the gripper for 10 steps, rises by $\min(\rho,0.05)$ for a retreat $\rho$ and, if $\rho\ge0.05$, parks the arm at $(-0.25,0,0.35)$, outside every view, before waiting 25 steps; holding keeps the gripper closed for 35 steps.
The cube slips by about 4\,mm in the grasp, so the nominal stack hover gaps of 1.4 and 2.4\,cm leave real gaps of about 1.0 and 1.9\,cm.

\paragraph{Design conditions.}
\Cref{tab:conditions} lists the conditions executed in every scene, in index order.
The classes are design intents; outcome and failure type are always re-read from the simulator.
StackCube has no stable overhang: releases up to 1.8\,cm off-centre stay on top and succeed, while from 2.1\,cm on the cube topples.
The preregistered class F-edge was therefore renamed F-fell; its cubes mostly end beside or behind the goal.
In tray, the same rim release can end in the tray, on the rim or outside it.

\begin{table*}[t]
  \centering\footnotesize
  \setlength{\tabcolsep}{5pt}
  \begin{tabular}{@{}lllllllll@{}}
    \toprule
    Task & Class & Indices & $S$ & Release offset (cm) & $h$ (cm) & Release & Retreat (cm) & Support\\
    \midrule
    stack & SUCC     & c00--c05 & 2, 2, 1 & $r=0,\ 0.7,\ 1.4$ along $\phi$ & 0.2 & yes & 2 or 10 & goal\\
          & F-hover  & c06--c09 & 2, 1    & $r=0,\ 1.6$ along $\phi$ & 1.4 or 2.4 & no & -- & goal\\
          & F-fell   & c10--c11 & 1       & $r=2.4,\ 3.0$ along $\phi$ & 0.4 & yes & 8 & goal\\
          & F-behind & c12--c13 & 2, 1    & $(\Delta x,\Delta y)=(-4.7,0),\ (-6.0,0)$ & 0.2 & yes & 8 & table\\
    \midrule
    tray  & SUCC     & c00--c05 & 2, 2, 1 & $r=0,\ 1.0,\ 2.0$ along $\phi$ & 0.4 & yes & 2 or 10 & goal\\
          & F-held   & c06--c09 & 2, 1    & $r=0,\ 1.5$ along $\phi$ & 3.0 or 5.0 & no & -- & goal\\
          & F-rim    & c10--c11 & 1       & $r=4.7,\ 5.5$ along $\phi$ & 2.4 & yes & 8 & goal\\
          & F-behind & c12--c13 & 2, 1    & $(\Delta x,\Delta y)=(-7.6,0),\ (-9.0,0)$ & 0.2 & yes & 8 & table\\
          & F-beside & c14      & 0       & $(\Delta x,\Delta y)=(0,\pm10)$, sign of $\sin\phi$ & 0.2 & yes & 8 & table\\
    \bottomrule
  \end{tabular}
  \caption{Design conditions per scene: 14 in stack, 15 in tray.
  $r$ is the release offset from the goal centre along the scene's direction $\phi$; $h$ is the nominal height of the cube bottom above the support (in tray the rim is 2.0\,cm above the floor, so F-held holds the cube 1 or 3\,cm above it); a retreat of 5\,cm or more parks the arm.
  $S$ (success-like appearance: 2 high, 1 mid, 0 low) is given per offset or distance, in the order listed.
  Indices run over the offsets and, within each, the near and far retreat (SUCC), and over the heights and, within each, the offsets (F-hover, F-held).}
  \label{tab:conditions}
\end{table*}

\paragraph{Outcome predicates.}
Let $p$ be the red cube's centre and $g$ the goal's position in the final state (the green cube's centre, or the tray's origin, at table level in its centre), $d=\lVert(p-g)_{xy}\rVert$, and $\delta_x=p_x-g_x$, $\delta_y=p_y-g_y$ along the world axes.
\emph{Success} in stack is the predicate of StackCube-v1~\cite{tao2024maniskill3}: $d\le\lVert(0.02,0.02)\rVert+0.005\approx0.033$, $|p_z-g_z-0.04|\le0.005$, and the cube is static and not grasped.
In tray, with $(l_x,l_y)$ the cube's horizontal offset in the tray's frame, success requires $\max(|l_x|,|l_y|)\le0.045-0.02+0.004=0.029$, $|p_z-g_z-0.025|\le0.006$ (resting on the floor), a static cube (ManiSkill's velocity test with thresholds $10^{-2}$ linear and $0.5$ angular) and no grasp.
Every other end state takes the first matching type:
\begin{enumerate}[nosep,leftmargin=*]
\item grasped: \emph{hover} if $b>0.006$, otherwise \emph{held\_on}, where $b=p_z-0.02-z_{\mathrm{s}}$ is the height of the cube bottom above the top $z_{\mathrm{s}}$ of the goal support (green-cube top or tray-floor top);
\item flat, \ie $|p_z-0.02|<0.006$ and a tilt below $10^\circ$ (the angle between world $z$ and the nearest cube axis), unless the task is tray and $d<0.045$: \emph{behind} if $\delta_x<0$ and $|\delta_y|<w$, with $w=0.03$ (stack) or $0.055$ (tray), otherwise \emph{beside};
\item $d<0.045$ (stack) or $d<0.06$ (tray): \emph{edge};
\item otherwise \emph{beside}.
\end{enumerate}
A cube lying on the tray floor also passes the height test of rule 2; the exception sends such a cube, if it is not a success, to rule 3.

\paragraph{Cue truth.}
\Cref{tab:probes} lists the four cue probes per task and the simulator predicates that define their correct answers, computed from the red cube's pose, the goal's pose and yaw, and the grasp flag.

\begin{table*}[t]
  \centering\footnotesize
  \setlength{\tabcolsep}{3pt}
  \begin{tabular}{@{}llp{0.36\linewidth}@{}}
    \toprule
    Cue & Question & Correct answer is Yes when\\
    \midrule
    \multicolumn{3}{@{}l}{\emph{stack}}\\
    contact   & Is the red cube touching the green cube? & $m<0.04$ and $|p_z-g_z-0.04|<0.003$; or on\_table holds and $m-0.04<0.003$\\
    grasped   & Is the robot gripper currently holding the red cube? & the grasp flag is set\\
    on\_top   & Is the red cube resting directly on top of the green cube? & $m<0.04$, $|p_z-g_z-0.04|<0.003$, and not grasped\\
    on\_table & Is the red cube resting on the table surface? & $|p_z-0.02|<0.004$ and tilt below $10^\circ$\\
    \midrule
    \multicolumn{3}{@{}l}{\emph{tray}}\\
    inside    & Is the red cube inside the blue tray? & $m<0.029$ and $p_z-0.02<0.023$\\
    grasped   & Is the robot gripper currently holding the red cube? & the grasp flag is set\\
    on\_floor & Is the red cube resting on the floor of the blue tray? & $m<0.029$, $|p_z-0.02-0.005|<0.003$, and not grasped\\
    on\_table & Is the red cube resting on the table surface outside the tray? & $|p_z-0.02|<0.004$, tilt below $10^\circ$, and $m\ge0.029$\\
    \bottomrule
  \end{tabular}
  \caption{Cue probes and their simulator truth. Each question is sent followed by ``Answer with Yes or No.''
  $m=\max(|l_x|,|l_y|)$, where $(l_x,l_y)$ is the red cube's horizontal offset from the goal in the goal's frame, rotated by the goal yaw.}
  \label{tab:probes}
\end{table*}

\paragraph{Views, lighting and rendering.}
\Cref{tab:views} gives the camera poses and lighting presets.
The cameras aim at the nominal goal, so the jittered goal lies up to 1\,cm per axis off their aim point.
Every end state is rendered from the four views as a final frame with segmentation masks, and as a 16-frame clip per view for the video judge: frames are taken at every fourth control step from the start of the descent to the release height to the end of the script, sampled uniformly to 16, resized to $384\times384$ and stored at 8\,fps.
Under the preregistered visibility table, a failure's counter-evidence is visible by design in every type--view cell except behind at V1 and hover at V3, and only if at least 300 pixels of the red cube are visible in the segmentation mask; in v1, the hand covers the cube so that fewer than 300 of its pixels show in 161 V3 stimuli.
Stimuli were generated on one machine and judged on another.
Re-rendering six end states (four views each) on the second machine changed pixel values by 0.03--0.10 grey levels on average (of 255) and by more than 10 levels in about $10^{-5}$ of the pixels, and segmentation counts by at most one pixel; re-simulating the same parameters there reproduced every failure type, with cube positions within 0.4\,mm.

\begin{table}[t]
  \centering\scriptsize
  \setlength{\tabcolsep}{2pt}
  \begin{tabular}{@{}llll@{}}
    \toprule
    View & Eye offset & Eye position & Image up\\
    \midrule
    V1 front-low  & $(0.42,0,0.10)$    & $(0.47,0,0.13)$     & $+z$\\
    V2 oblique    & $(0.30,0.30,0.30)$ & $(0.35,0.30,0.33)$  & $+z$\\
    V3 top-down   & $(0,0,0.50)$       & $(0.05,0,0.53)$     & $-x$\\
    V4 level side & $(0,0.42,0.035)$   & $(0.05,0.42,0.065)$ & $+z$\\
    \midrule
    Lighting & Ambient & Key light: direction, intensity & Fill\\
    \midrule
    preset 0 & 0.30 & $(1,1,-1)$, 1.0    & 1.0\\
    preset 1 & 0.45 & $(-0.6,1,-1)$, 0.8 & 0.6\\
    \bottomrule
  \end{tabular}
  \caption{Cameras and lighting presets. Every camera looks at $(0.05,0,0.03)$, 3\,cm above the nominal goal centre; the eye offset is relative to that point, the eye position in world coordinates (m). All cameras render $512\times512$ pixels with a field of view of 0.45\,rad and clipping planes at 0.01 and 100\,m. Lights are directional and grey; the key light casts shadows (shadow map $2048\times2048$), and the fill light points straight down, $(0,0,-1)$.}
  \label{tab:views}
\end{table}

\paragraph{Natural distribution $\pi_0$.}
For scene seed $s$, $\theta\sim\pi_0$ comes from a generator seeded with $50{,}000+s$, which draws its own direction $\phi\sim\mathcal{U}(0,2\pi)$; $\pi_0$ was fixed in amendment v1-2, before any judge data.
\begin{itemize}[nosep,leftmargin=*]
\item With probability 0.1 the cube is placed on the table ($h=0.002$, release, retreat 0.10): with probability $\frac12$ behind the goal, $(\Delta x,\Delta y)=(-b,u)$ with $b\sim\mathcal{U}(0.045,0.10)$ in stack or $\mathcal{U}(0.07,0.12)$ in tray and $u\sim\mathcal{U}(-0.03,0.03)$; otherwise beside it, $(\Delta x,\Delta y)=(u,\pm c)$ with $c\sim\mathcal{U}(0.06,0.12)$ and either sign with probability $\frac12$.
\item Otherwise the controller aims at the goal: $r=|n|$ with $n\sim\mathcal{N}(0,0.012^2)$ with probability 0.85 and $r\sim\mathcal{U}(0.02,0.06)$ otherwise, and $(\Delta x,\Delta y)=r(\cos\phi,\sin\phi)$; it releases with probability 0.8; $h\sim\mathcal{U}(0.002,0.03)$ in stack or $\mathcal{U}(0.004,0.05)$ in tray; the retreat is 0.02 or 0.10 with probability $\frac12$ each.
\end{itemize}

\begin{figure*}[t]
  \centering
  \includegraphics[width=\textwidth]{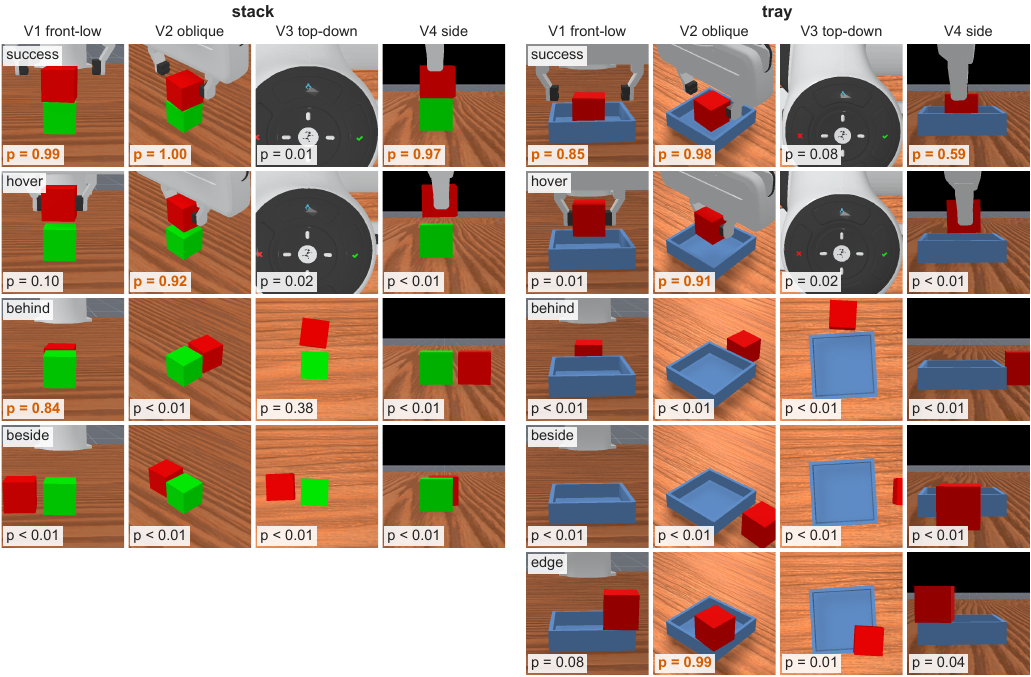}
  \caption{Stimulus catalogue: one main-set scene per task (stack seed 1, tray seed 6) and, in it, one end state per design class, each rendered from the four views (centre crops); labels give the simulator's outcome type.
  Numbers: M1's success probability $p$, in bold where M1 accepts ($p\ge0.5$).
  These examples illustrate view-dependent evidence, rather than population error rates.}
  \label{fig:catalog}
\end{figure*}

\paragraph{Stimulus sets.}
\Cref{tab:sets} lists the sets and their simulator outcomes.
All share the scene randomisation above; v1 and v2 execute the design conditions, N1 and N2 one draw $\theta\sim\pi_0$ per scene.
Every end state is rendered from the four views, so each set has four times as many stimuli as end states, and half of the scenes of each set use each lighting preset.
The two tasks use the same seed numbers; seed ranges never overlap across sets, and the closed-loop rollouts use seeds from 200{,}000 on (\cref{sec:testbed}).
\Cref{fig:catalog} shows one end state per design class in one scene of each task.

\begin{table*}[t]
  \centering\footnotesize
  \setlength{\tabcolsep}{5pt}
  \begin{tabular}{@{}lllcrrrrrrr@{}}
    \toprule
    Set & Scene seeds & Use & Task & End states & success & hover & held\_on & behind & beside & edge\\
    \midrule
    v1 & 0--11    & diagnosis      & stack & 168 & 73  & 48 & 0  & 32 & 15 & 0\\
       &          &                & tray  & 180 & 76  & 48 & 0  & 28 & 16 & 12\\
    v2 & 20--31   & replication    & stack & 168 & 73  & 48 & 0  & 24 & 23 & 0\\
       &          &                & tray  & 180 & 80  & 48 & 0  & 24 & 19 & 9\\
    N1 & 100--199 & calibration    & stack & 100 & 57  & 10 & 11 & 9  & 13 & 0\\
       &          &                & tray  & 100 & 64  & 20 & 1  & 5  & 6  & 4\\
    N2 & 200--499 & H3 search pool & stack & 300 & 170 & 37 & 15 & 22 & 54 & 2\\
       &          &                & tray  & 300 & 207 & 47 & 5  & 10 & 21 & 10\\
    \bottomrule
  \end{tabular}
  \caption{Stimulus sets and simulator outcomes per task. In v1, seeds 0--7 form the main set (112 stack and 120 tray end states) and seeds 8--11 are held out. Stimuli per set: 1,392 in v1 and in v2, 800 in N1, 2,400 in N2.}
  \label{tab:sets}
\end{table*}

\section{Scoring and Closed-Loop Implementation}
\label{sec:supp_closed}
\paragraph{Success prompt.}
The single-frame judges receive: ``The robot was given this instruction: `\textless instruction\textgreater'. Look at the final scene. Did the robot successfully complete the instruction? Answer with Yes or No.'' The task instructions are ``Stack the red cube on top of the green cube.'' and ``Put the red cube into the tray.'' Each request is a single user turn with one image and one question, without a system prompt or prior conversation. Separate cue requests append ``Answer with Yes or No.'' to the questions in \cref{tab:probes}. The cue regression is descriptive logistic regression on a judge's four own probe responses, with 1,000 scene-bootstrap replicates. Unparsed probe responses enter as $.5$.

\paragraph{Closed-model settings.}
Gemini 3.6-flash uses the Gemini API, minimal reasoning, temperature 0, 1,024 maximum output tokens, and no seed. Claude Opus 5.5 uses Anthropic Message Batches initially, low effort, 2,000 maximum tokens, and no sampling parameters; 232 of its 1,200 N2 verdicts later use the pinned Anthropic route through OpenRouter. Gemini 3.8-flash, GPT-6 Astra, and Claude Fable 5.1 use OpenRouter with the original vendor pinned and fallback disabled. Gemini uses minimal reasoning, temperature 0, seed 0, and 1,024 maximum tokens; Astra uses minimal reasoning, seed 0, and 2,000 maximum tokens; Fable uses low effort and 2,000 maximum tokens. These are historical experiment settings, not recommendations for present endpoints. The archived identifiers are \texttt{gemini-3.6-flash}, \texttt{claude-opus-5-5}, and, on OpenRouter, \texttt{anthropic/claude-opus-5.5}, \texttt{google/gemini-3.8-flash}, \texttt{openai/gpt-6-astra}, and \texttt{anthropic/claude-fable-5.1}. Reported version strings include \texttt{gemini-3.8-flash-20260902} and \texttt{claude-fable-5.1-20260831}.

An answer is stripped of whitespace and leading formatting marks; case-insensitive initial Yes or No maps to 1 or 0. Anything else is unparsed and maps to $.5$. The three later OpenRouter-only judges re-ask once after an unparsed answer. Earlier Gemini and Opus data have no re-ask. Fable abstains on 63/1,392 diagnostic verdicts and 134/2,400 N2 verdicts; Gemini 3.6 on 15 and 24; Gemini 3.8 on 2 and 3. Opus and Astra have no unparsed verdicts. Reclassifying abstentions as rejections in N2 only, or in both diagnosis and N2, changes no H3 pass/fail verdict. The original Opus diagnostic coverage is 696 V1/V2 images, not 1,392 four-view requests.

\paragraph{Routes and incomplete coverage.}
Unpinned route checks failed their original agreement threshold and their data were not merged. With provider pinning, Opus agreed on 58/60 previously scored rows and Gemini 3.6 on 57/60; 30-row same-route retests agreed on 27 and 28. These limited checks do not establish route equivalence. Only Opus actually merges routes, with 968 original and 232 pinned-route N2 rows; excluding the latter does not change H3 under the pre-fixed subset rule. Gemini 3.6's secondary diagnostic replication stopped at 125/1,392 rows under rate limits and was not completed. Its primary diagnosis and N2 are complete. Reduced concurrency, same-provider resumptions, and retry-on-empty-response changes are operational deviations, not additional scientific observations. Optional increased-reasoning controls were not run.

\paragraph{Other judges.}
J1 and M1 are frozen open checkpoints, scored with normalized Yes/No probabilities. M1's archived model revision is \texttt{1d4bf0f}; its parameters and generation differ from J1, so their comparison cannot isolate scale. RoboReward receives uniformly sampled 16-frame placement clips at $384\times384$, with a threshold calibrated on N1 for its expected 1--5 progress score. J4 trains a logistic-regression head on frozen DINOv2 features using N1, at V2 only. Their different inputs prevent interpreting the nine judges as an equal-information model ranking. Only J1/M1 participate in the completed multi-seed loops.

\paragraph{Refitting details.}
For a Bernoulli branch with $k$ positive decisions among $n$ eligible retained samples, the implemented estimate is $\operatorname{clip}((k+1)/(n+2),.02,.98)$; an empty eligible branch gives $.5$. Table probability uses all retained samples, release probability uses goal-branch samples, and far-retreat probability uses released goal-branch samples. Goal and table offsets use empirical means and sample covariance plus $10^{-6}I$ when more than two samples exist, otherwise covariance $10^{-4}I$. If a branch has no observations, its previous continuous parameters persist. Log heights use $\log\max(h,.001)$, their empirical mean, and population standard deviation floored at $.05$. Sampling clips the exponentiated normal to $[.002,.05]$ metres. The table branch always releases at 2\,mm and retreats far; the goal branch samples the release/retreat decisions. Thus add-one smoothing, covariance regularization, and clipping mean the update is not a pure unconstrained maximum-likelihood operator.

\paragraph{Frozen cell definitions and repairs.}
The dominant type maximizes main-set failure acceptance among types with at least four states. The target set adds other eligible types whose acceptance is at least $.20$. C6 takes the smallest threshold with at most 5\% dominant-type acceptance and uses a strict $p>\tau_c$ test. CV chooses the other view with the lowest dominant-type acceptance among views retaining at least 75\% of successes, and queries it only after the deployment view accepts. C2 uses N1 TPR/FPR, including held\_on. C3t matches the retained count of CV in inseparable cells and C6 otherwise, selecting the top scores. These definitions were fixed before the historical loop.
\begin{table}[ht]
  \centering
  \footnotesize
  \setlength{\tabcolsep}{2.6pt}
  \begin{tabular}{@{}llccll@{}}
    \toprule
    Cell & Class & AUROC & $\tau_c$ & $\mathcal{B}_c$ & $v'_c$\\
    \midrule
    J1/stack/V2 & separable     & .988 & .998 & hover         & V4\\
    M1/stack/V2 & separable     & .970 & .984 & hover         & V4\\
    M1/tray/V2  & separable     & .980 & .915 & hover, edge   & V1\\
    J1/tray/V1  & inseparable   & .485 & .993 & behind        & V4\\
    J1/tray/V2  & inseparable   & .820 & .993 & edge, hover   & V1\\
    M1/stack/V1 & inseparable   & .652 & .967 & behind        & V2\\
    J1/stack/V1 & intermediate  & .940 & .995 & behind, hover & V2\\
    M1/tray/V1  & no blind spot & --   & .223 & --            & V4\\
    \bottomrule
  \end{tabular}
  \caption{Frozen cell parameters, computed from the v1 main set before any loop data.
  AUROC separates true successes from the dominant type $f^\star_c$ (listed first in $\mathcal{B}_c$) at the deployment view; $\tau_c$ is the C6 threshold and $v'_c$ the CV check view.
  No failure type passes in M1/tray/V1; its $\tau_c$ and $v'_c$ are computed for hover.}
  \label{tab:cells}
\end{table}

\section{Prespecification, Deviations, and Evidence Scope}
\label{sec:supp_prereg}
The term preregistered refers to local dated or hash-locked study documents, not registration with a public service. Early amendments v1--v1-5 were dated before judge data; v1-6 onward had adjacent SHA-256 locks. Per-judge H3 forecasts were frozen before N2 scoring, and the loop analysis was fixed before loop data. File dates and hash locks document a workflow, not an independent third-party timestamp. The supplied evidence index preserves the original files and hashes.
\begin{table*}[t]
\centering\footnotesize\setlength{\tabcolsep}{4pt}
\begin{tabular}{@{}p{.26\textwidth}p{.24\textwidth}p{.24\textwidth}p{.18\textwidth}@{}}\toprule
Claim or result & Evidence & Prespecification & Replication unit\\\midrule
Type/view dependence; cue use & Truth-fixed diagnosis; H3 & Dated/amended protocols; later H3 intervals & Scenes; 12/task in diagnosis, 300/task in N2\\
Missing below-rim support & Historical F1 D/D+ & Hash-locked before F1 data & 40 paired valid scenes\\
Fewer but harder residual hovers & Revised N0/R0/C1 & Revised after historical observations; reuses F1 seed IDs & 20 paired seeds; pooled rounds\\
State-specific filtering effects & Revised C3/C2* & Later protocol, frozen before its run & 5 paired seeds per cell\\
Retention and safeguard outcomes & Historical T2--T4 & Frozen before main loop; deviations retained & 5 seeds per cell\\
Static-ratio prediction fails & Later ratio/trajectory audit & Exploratory, post hoc & 8 related cells\\\bottomrule
\end{tabular}
\caption{Evidence scope at a glance. Prespecification refers to local dated or hash-locked documents. Revised and historical protocols are analyzed separately; arm-round records are not independent replicates.}
\label{tab:evidence_scope}
\end{table*}

\begin{table*}[t]
  \centering\footnotesize
  \setlength{\tabcolsep}{6pt}
  \begin{tabular}{@{}lll@{}}
    \toprule
    Document & Frozen & SHA-256 (prefix) / content\\
    \midrule
    Diagnosis v1 & 10-02 & hypotheses H1--H3, stimulus design\\
    v1-1 to v1-5 & 10-02 & visibility table, pools N1/N2, cue analysis, J2 threshold\\
    v1-6 & 10-03 & \texttt{0a41f9a5} mid-size judge M1\\
    Pilot v2-0 & 10-03 & \texttt{cfa29ad7} closed-loop pilot\\
    v1-7 & 10-03 & \texttt{283f23cd} first closed judges\\
    v1-8 & 10-05 & \texttt{531595d9} Claude batch route\\
    v1-9 & 10-05 & \texttt{4c599e49} OpenRouter route\\
    Main experiment v3 & 10-05 & \texttt{52b7b048} closed loop, T1--T4\\
    v1-10 & 10-05 & \texttt{03ce92eb} pinned providers, strongest models\\
    v1-11 & 10-06 & \texttt{cb24660e} scope, abstention rule, budget\\
    Follow-up F1 & 10-06 & \texttt{8cf44880} M1/tray/V1 follow-up, F1a--F1c\\
    \bottomrule
  \end{tabular}
  \caption{Preregistration documents (dates in 2026).}
  \label{tab:prereg}
\end{table*}

\paragraph{Original decision rules.}
H3 requires $\rho_S\ge .5$ and lower mean KL than B1 and B2. T1 applies that same rule to C3's accepted failures, pooling rounds and seeds, over cells with at least five accepted failures. T2 requires positive C3--C1 target-failure differences with seed-bootstrap intervals excluding zero and point differences exceeding C2--C1 in at least five of seven blind-spot cells. It passes in six. T3 requires two of three cells in each separability class to pass: in separable cells, C6 must be superior to C3 and non-inferior to C3t (one-sided 95\% bound above $-.02$); in inseparable cells, CV must beat C6 and C3t and have the least blind-spot contamination. T3 fails overall. T4 uses jointly resampled seeds and a $-.05$ non-inferiority margin for M1 minus J1 blind-spot share among accepted failures. It passes only that restricted composition criterion. Historical bootstrap intervals use 10,000 seed resamples and are not familywise-adjusted.

\paragraph{Deviations that affect interpretation.}
The original H3 protocol also described top-5\% selection with one local refinement and a scene-bootstrap KL comparison. Refinement was not run; the original verdict used point estimates. The scene bootstrap was implemented later and is reported as supplementary uncertainty, not retroactive confirmation. J4 was initially counted in an internal passing summary; that count is corrected because B2 equals its map. The T4 frozen script includes held\_on in its denominator despite the general error-map exclusion; both denominator choices are reported. C3t duplicates C6 whenever its count is matched to C6, so non-inferiority is uninformative there. The parameter file's separable label for M1/tray/V1 is superseded by the prespecified no-blind-spot rule. Loop rendering uses one lighting preset whereas diagnostic scenes alternate two. A historical single-seed diffusion pilot was not expanded into the planned multi-seed secondary testbed. None of these deviations is silently recoded as a successful test.

\paragraph{Later analyses.}
The unified N0/R0/C1 reset, C2*, the acceptance-ratio study, and the detailed implementation audit followed observation of earlier results. Their protocols may be frozen before their own execution, but they are not original prespecified tests of the current storyline. The central 20-seed mechanism run reuses F1 seed numbers. Its C1--R0 below-rim contrast has a positive one-sided 95\% lower bound of $.0063$, but its two-sided interval $[-.0069,.1455]$ crosses zero. Both facts are retained. Later discussion was informed by CriticHack and related verifier-error literature; chronology alone does not establish independent conceptual priority.

\paragraph{Completed versus pending.}
The revised R0, M1, and J1 runs contribute 92,000 formal arm-round records, with complete archived delivery. A separate subsequent pilot contributes 4,500 pilot armrows from 3,300 physical rollouts; these are excluded from formal evidence. Twelve additional configurations have frozen scientific code but no completed formal run in the latest project record read for this draft. The planned re-scoring of 2,220 historical C1 failures has zero completed scores. Consequently the truth-filtered monitoring result is stated for one controlled setting, not all eight.

\section{Static Prediction and Its Uncertainty}
\label{sec:supp_h3}
For each judge, the correlation concatenates task--view--type entries from task--view pairs with at least one accepted N2 failure. KL averages over those eligible pairs. Additive smoothing of $10^{-3}$ per type precedes normalization. Sparse error cells can become empty in bootstrap replicates and follow the original estimator's exclusion rule. The later uncertainty analysis resamples scenes 2,000 times, retaining within-scene multi-view dependence. Its primary fixed-forecast variant holds the diagnostic map and forecast fixed; a second variant reweights predictions using resampled pool counts. Neither gives a judge positive lower bounds against both baselines. This is uncertainty about advantage, not proof that the map contains no information.

\begin{table*}[t]
\centering\small\setlength{\tabcolsep}{5pt}
\begin{tabular}{@{}lrrrrll@{}}\toprule
Judge & $\rho_S$ & Map KL & B1 KL & B2 KL & B1$-$map [95\% CI] & B2$-$map [95\% CI]\\\midrule
J1 Qwen3-VL-8B & .707 & .560 & 1.124 & 2.133 & $.563\ [-.043,.908]$ & $1.573\ [.347,1.934]$\\
M1 Qwen3.8-27B & .582 & .702 & .919 & 1.245 & $.217\ [-.094,.514]$ & $.544\ [-.366,1.543]$\\
RoboReward-8B & .684 & 1.061 & 1.116 & 2.031 & $.055\ [-.970,.788]$ & $.970\ [-1.262,1.990]$\\
Claude Opus 5.5 & .583 & .387 & 1.099 & .477 & $.712\ [.230,1.033]$ & $.090\ [-.526,.469]$\\
Gemini 3.6-flash & .451 & 1.616 & 1.152 & .708 & $-.464\ [-1.409,.159]$ & $-.908\ [-1.879,-.272]$\\
Gemini 3.8-flash & .446 & 1.255 & 1.832 & 1.190 & $.577\ [-.530,.935]$ & $-.066\ [-1.196,.215]$\\
Claude Fable 5.1 & .434 & 1.133 & .748 & .998 & $-.384\ [-.867,.088]$ & $-.135\ [-.896,.364]$\\
GPT-6 Astra & .515 & .541 & 1.294 & .727 & $.754\ [.136,1.026]$ & $.187\ [-.511,.501]$\\\bottomrule
\end{tabular}
\caption{Complete H3 point estimates and later fixed-forecast scene-bootstrap intervals. Advantage is baseline KL minus map KL; positive favors the map. Rounding can make differences of displayed KLs differ from displayed advantages. J4 is excluded from the eight-way pass count: $\rho=.894$, map/B2 KL $.637$, B1 KL $1.204$, at V2 only. Intervals are unadjusted.}
\label{tab:h3}
\end{table*}

\paragraph{T1 sensitivity.}
M1/tray/V1 has an all-zero pre-smoothing forecast, hence a uniform smoothed forecast. At $\epsilon=10^{-3}$ its KL is $.884$. Omitting the cell gives mean KL $.311$ versus B1 $.360$ and B2 $.898$; replacing its forecast by B1 gives $.299$ versus $.342$ and $.792$. The favorable inequality holds for these post hoc variants over $\epsilon=10^{-6}$ to $10^{-2}$, but all variants fail at $.05$. The registered test includes the cell and fails. Thus both uncovered states and the convention for zero predicted acceptance affect the aggregate measure.

\section{Original F1 and the Revised Reset Run}
\label{sec:supp_t1}
The following F1 results are the original registered follow-up, retained for transparency. They must not be pooled with the revised N0/R0/C1 data in the main text. In particular, F1 N0 has 490 accepted hovers and F1 C1 has 176 hovers, whereas the revised run has 497 and 175, respectively.
\paragraph{Follow-up F1: design.}
The follow-up was hash-locked on 10-06 (\cref{tab:prereg}), before its data; a short preliminary test was excluded.
It repeats cell M1/tray/V1 with new policy seeds (5--24) and new scenes, 100 rollouts per arm and round for six rounds, and the judge scores every rollout of every arm.
All arms use the same scene for the same seed, round and rollout; C1 and C3 draw their parameters with common random numbers and are refit on the rollouts they accept, while N0 draws fresh parameters from $\pi_0$ and is never fitted.
D contains the original above-rim hovers (target heights 3 and 5\,cm above the tray floor, measured about 2.6 and 4.55\,cm), and D+ the same scenes at target heights 1.5 and 2.0\,cm (measured about 1.0 and 1.55\,cm), each at gripper offsets of 0 and 1.5\,cm, in 40 scenes at one lighting preset.
A scene counts only if all eight of its stimuli are hovers, those of D above the rim and those of D+ below it.
Only rounds 1--5 enter the tests, and a seed counts for an arm only if that arm has at least three hovers there.
F1a compares the one-sided 97.5\% $t$ lower bound of the mean of C3's per-seed acceptance rates with the one-sided 97.5\% Clopper--Pearson upper bound of the share of D scenes with at least one pass; as each scene has exactly four D stimuli, this bound is not below the mean D acceptance whatever the correlation within a scene.
F1b requires one-sided 95\% $t$ lower bounds above 0 for the per-seed differences C1$-$N0 in both the acceptance of hovers and their share below the rim.
F1c, tested only if both hold, requires a one-sided 95\% $t$ lower bound above 0 for the per-scene difference in acceptance between D+ and D.
A test is undecided with fewer than 12 seeds or 30 scenes.
In a power simulation with parameters from the main experiment, boundary nulls passed F1a at a rate of 0.000, F1b at most 0.049 and F1c at most 0.053.
The run took 4.0\,h on one GPU, and all 20 seeds completed at their first attempt.

\begin{figure*}[t]
  \centering
  \includegraphics[width=\textwidth]{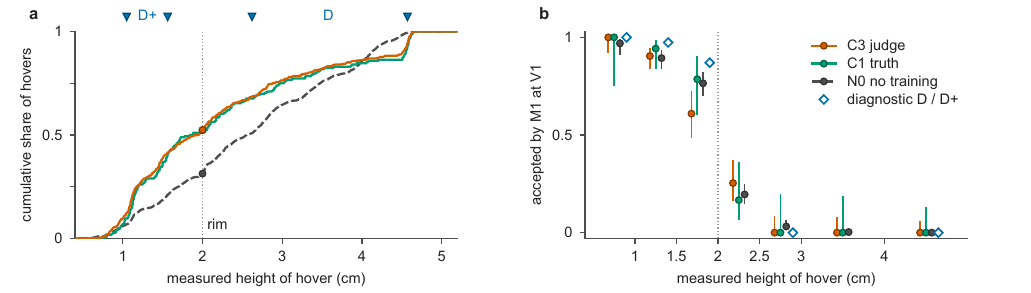}
  \caption{Follow-up F1 (M1/tray/V1; rounds 1--5 of 20 seeds).
  (a) Cumulative distribution of the measured heights of hovers per arm; dots: share below the tray rim (2\,cm); triangles: measured heights of the diagnostic hovers, two design heights each for D+ and D.
  (b) Share of hovers accepted by M1 at V1 per height bin (bins as in \cref{tab:f1}; Wilson 95\% intervals), and of the diagnostic stimuli of D and D+ (diamonds).}
  \label{fig:f1}
\end{figure*}

\paragraph{Follow-up F1: further results (descriptive).}
\Cref{fig:f1} and \cref{tab:f1} give the distributions and counts.
Acceptance has a similar coarse dependence on measured height across arms, but conditional response equivalence was not tested; per seed, acceptance under the judge and under truth filtering did not differ (C3$-$C1: $-0.025$ $[-0.088,0.038]$), and C3 exceeded N0 by $0.173$ $[0.130,0.216]$.
The blind spot depends on the camera: scored at the other views, D+ passed in none of 160 stimuli at the top camera, in 16 at the side camera and in all 160 at the oblique one (D: 0, 0 and 85).

\begin{table*}[t]
  \centering\footnotesize
  \setlength{\tabcolsep}{5pt}
  \begin{tabular}{@{}lrrrrrrr@{}}
    \toprule
    & & & & \multicolumn{4}{c}{accepted / hovers, by measured height}\\
    \cmidrule(l){5-8}
    Set & hovers & below rim & accepted & $<1.5$\,cm & 1.5--2.0\,cm & 2.0--2.5\,cm & $\ge 2.5$\,cm\\
    \midrule
    Round 0 (shared) & 322 & 96 & 96 & 57/62 & 25/34 & 14/68 & 0/158\\
    C3, rounds 1--5 & 469 & 246 & 225 & 169/182 & 39/64 & 17/67 & 0/156\\
    C1, rounds 1--5 & 176 & 92 & 87 & 61/64 & 22/28 & 4/24 & 0/60\\
    N0, rounds 1--5 & 1592 & 499 & 490 & 271/295 & 156/204 & 54/276 & 9/817\\
    \midrule
    D & 160 & 0 & 0 & & & & 0/160\\
    D+ & 160 & 160 & 148 & 81/83 & 67/77 & & \\
    \bottomrule
  \end{tabular}
  \caption{Follow-up F1 (M1/tray/V1; 20 seeds, 2,000 rollouts per arm and round; diagnostic sets at V1): hovers, hovers below the rim (measured height below 2\,cm) and hovers accepted by M1.
  The true success rate in round~5 was 0.963 under C3, 0.975 under C1 and 0.682 under N0, against 0.663 in round~0.}
  \label{tab:f1}
\end{table*}

\paragraph{Historical discrepancies.}
The audit of original F1 versus common-reset round 0 finds 12 judgement/type disagreements from 11 seeds: six truth changes and six judgement-threshold-only changes. The original equivalence gate did not pass; it remains a failed historical check. The revised protocol jointly reruns all three arms and passes its own reset and worker-consistency checks. That does not erase the earlier mismatch. The C2* J1 historical audit similarly records five tasks and seven score discrepancies despite matching physical truth and type. Some differences are near $.5$, others larger. Current-protocol reproducibility and historical reproducibility are distinct claims.

\paragraph{Revised-run denominators.}
Each seed contributes 500 observations per arm over rounds 1--5. A conditional ratio is eligible only when the arm has at least three hovers in this pooled window; all 20 seeds qualify. C1 has 4--15 hovers per seed and 14 zero-hover seed-rounds, which are not selectively discarded: rounds are pooled before dividing. Equal-seed hover acceptance is 49.98\%; pooled acceptance is $86/175=49.14\%$. No uncertainty calculation treats 10,000 arm rows as independent trials. The main-text height-stratum counts are descriptive pooled numerators and denominators without binomial independence assumptions.

\begin{table*}[t]
\centering\footnotesize\setlength{\tabcolsep}{6pt}
\begin{tabular}{@{}r|rrrr|rrrr|rrrr@{}}\toprule
&\multicolumn{4}{c|}{N0}&\multicolumn{4}{c|}{R0}&\multicolumn{4}{c}{C1}\\
Seed & $H$ & $A$ & $L$ & $S$ (\%) & $H$ & $A$ & $L$ & $S$ (\%) & $H$ & $A$ & $L$ & $S$ (\%)\\\midrule
5 & 80 & 22 & 24 & 69.6 & 62 & 24 & 22 & 68.6 & 9 & 4 & 4 & 96.2 \\
6 & 70 & 26 & 25 & 67.8 & 83 & 47 & 42 & 66.4 & 10 & 6 & 7 & 96.4 \\
7 & 71 & 18 & 19 & 69.0 & 114 & 60 & 59 & 63.6 & 8 & 4 & 4 & 96.2 \\
8 & 90 & 31 & 32 & 64.8 & 104 & 42 & 50 & 65.8 & 6 & 4 & 5 & 97.6 \\
9 & 70 & 16 & 18 & 67.6 & 70 & 30 & 36 & 72.0 & 9 & 4 & 4 & 98.2 \\
10 & 68 & 21 & 24 & 67.2 & 91 & 29 & 37 & 70.2 & 11 & 5 & 5 & 96.0 \\
11 & 71 & 24 & 23 & 70.2 & 208 & 68 & 68 & 43.8 & 12 & 4 & 4 & 97.4 \\
12 & 92 & 23 & 20 & 63.2 & 85 & 37 & 45 & 61.6 & 6 & 3 & 3 & 96.8 \\
13 & 81 & 32 & 29 & 65.8 & 93 & 33 & 33 & 68.8 & 11 & 3 & 3 & 97.0 \\
14 & 84 & 21 & 27 & 65.0 & 46 & 14 & 19 & 77.0 & 4 & 1 & 1 & 98.8 \\
15 & 71 & 16 & 14 & 69.4 & 72 & 34 & 39 & 72.8 & 6 & 5 & 5 & 97.6 \\
16 & 82 & 25 & 28 & 64.0 & 93 & 43 & 52 & 64.4 & 10 & 8 & 8 & 96.4 \\
17 & 77 & 22 & 21 & 69.2 & 63 & 33 & 34 & 72.2 & 10 & 4 & 4 & 98.0 \\
18 & 74 & 18 & 20 & 68.4 & 43 & 21 & 21 & 79.6 & 7 & 4 & 3 & 96.6 \\
19 & 83 & 34 & 32 & 66.0 & 70 & 22 & 24 & 68.6 & 5 & 2 & 2 & 98.6 \\
20 & 80 & 33 & 28 & 66.6 & 73 & 36 & 39 & 66.2 & 7 & 5 & 6 & 98.2 \\
21 & 83 & 30 & 30 & 67.2 & 104 & 42 & 51 & 64.2 & 9 & 7 & 7 & 96.8 \\
22 & 81 & 27 & 27 & 66.6 & 120 & 50 & 58 & 65.6 & 15 & 8 & 9 & 94.8 \\
23 & 85 & 25 & 26 & 62.6 & 68 & 21 & 23 & 66.4 & 10 & 2 & 4 & 96.6 \\
24 & 99 & 33 & 32 & 66.0 & 102 & 35 & 43 & 56.0 & 10 & 3 & 3 & 95.6 \\

\bottomrule\end{tabular}
\caption{Revised mechanism run, all 20 seeds. $H$: hover count; $A$: accepted hovers; $L$: hovers below the rim; $S$: true success percentage. Every arm has 500 observations per seed. These counts reproduce both seed means and pooled ratios.}
\label{tab:r0seeds}
\end{table*}

\section{Filtering Identities and the Role of Projection}
\label{sec:supp_math}
\paragraph{Exact one-step identity.}
Let $z$ be a resampleable controller parameter (or another precisely defined object) with current density $p_t(z)$. Let $\alpha_t(z)$ be its acceptance probability after integrating execution and scene randomness. Before refitting, selected samples have density
\begin{equation}
 q_t(z)=\frac{p_t(z)\alpha_t(z)}{Z_t},\qquad Z_t=\int p_t(z)\alpha_t(z)\,dz.
\end{equation}
For measurable sets $B,S$ with positive probabilities and mean acceptances $\bar\alpha_{B,t},\bar\alpha_{S,t}$,
\begin{equation}
 \frac{q_t(B)}{q_t(S)}=\frac{p_t(B)}{p_t(S)}\frac{\bar\alpha_{B,t}}{\bar\alpha_{S,t}}.
\end{equation}
The failure/success ratio increases when $\bar\alpha_{B,t}>\bar\alpha_{S,t}$, whereas the \emph{absolute} selected probability of $B$ increases when $\bar\alpha_{B,t}>Z_t$. These are different conditions. In an outcome-level joint distribution the same conditional-selection identity holds, but one cannot automatically treat the selected outcome distribution as a directly resampleable policy.

Only if fitting exactly reproduces $q_t$, the acceptance function is fixed, and its support remains appropriate does iteration give $p_T(z)\propto p_0(z)\alpha(z)^T$. For a deterministic accept/reject function on a fixed image, $\alpha\in\{0,1\}$ and $\alpha^T=\alpha$ for $T\ge1$: repeated powers alone provide no continuing dynamics. Our executions regenerate states, and our fitting is restricted and smoothed. The appropriate descriptive operator is $p_{t+1}=\operatorname{Fit}_{\mathcal P}(q_t)$, with finite-sample and numerical effects. CriticHack's exponential tilt under reward optimization~\cite{luo2026critichack} is not the same update as threshold-filtered imitation.

\paragraph{Why R0 is not an identity operation.}
For a tray goal branch, the initial command height is $H\sim U(a,b)$ with $a=.004$, $b=.05$ metres. The population log-normal projection matching log moments has
\begin{align}
 \mu&=\frac{[h\log h-h]_a^b}{b-a},\\
 \sigma^2&=\frac{[h\{(\log h)^2-2\log h+2\}]_a^b}{b-a}-\mu^2.
\end{align}
This gives $\mu=-3.77610$, $\sigma=.63011$, and median $e^\mu=22.91$\,mm. The probability of a command below 20\,mm changes from $.34783$ under the uniform to $.41461$ under this ideal log-normal projection; clipping at 2 and 50\,mm does not alter that interior-threshold probability. This calculation is an analytic illustration made during the audit, not a newly simulated result or a prediction of measured cube height. Grasp slip separates command height from actual bottom height. Finite-sample fitting, probability smoothing, variance constraints, and later rounds further separate it from R0.

Archived R0 command-height summaries support both a mean shift and narrowing: means across seeds are 26.87, 25.66, and 23.83\,mm at rounds 0, 1, and 5; corresponding mean within-seed standard deviations are 13.22, 13.21, and 10.63\,mm. These observations do not isolate projection from recursive fitting. All 100 post-initial C1 seed-rounds reach the $.98$ release cap. A shared height distribution and a nonzero holding branch can generate low residual hovers even though C1 selects no failed sample. Separate-branch, no-cap, and matched-family formal interventions remain uncompleted.

\section{Acceptance-Ratio Prediction: Negative Results}
\label{sec:supp_ratio}
The post hoc ratio is the static target-type acceptance rate divided by static true-success acceptance. It uses the original main diagnostic split and the frozen target sets. In six cells the success acceptance estimate is one, so the ratio reduces to target acceptance. Sample proportions such as $0/32$ or $17/17$ are not assertions that the population probabilities are exactly zero or one.

\begin{table}[t]
\centering\footnotesize\setlength{\tabcolsep}{4pt}
\begin{tabular}{@{}lrrr@{}}\toprule
Cell & Success accepted & Target accepted & Ratio\\\midrule
M1 / stack / V1 & 42/49 & 11/18 & .713\\
M1 / stack / V2 & 49/49 & 17/32 & .531\\
M1 / tray / V1 & 51/51 & 0/32 & .000\\
M1 / tray / V2 & 51/51 & 18/40 & .450\\
J1 / stack / V1 & 49/49 & 24/50 & .480\\
J1 / stack / V2 & 49/49 & 19/32 & .594\\
J1 / tray / V1 & 51/51 & 17/17 & 1.000\\
J1 / tray / V2 & 49/51 & 12/40 & .312\\\bottomrule
\end{tabular}
\caption{Static ratio inputs. The outcome is the continuous C3--C2* final target-failure difference in \cref{tab:c2star}, not a significance-based effective/ineffective label.}
\end{table}

Across these eight related cells, Spearman correlations with the final effect are $.407$ for the ratio, $.623$ for static FPR, $-.343$ for static accuracy, and $.455$ for target minus overall failure acceptance. The equal $0.2\pp$ effects are tied using integer final counts; an earlier floating-point rank split was corrected and retained in the audit. No correlation difference is tested as a model-ranking claim.

The trajectory approximation multiplies initial type frequencies by fixed type acceptance to the round power, then normalizes. Missing diagnostic types fall back to overall failure acceptance as recorded in the analysis. MAE averages equally over eight cells, five seeds, and six subsequent rounds; those 240 terms are not independent samples. It is retrospective and conditional on the actual round-0 distribution, without fitting an exponent to final outcomes.

\begin{table}[t]
\centering\footnotesize
\begin{tabular}{@{}lrrr@{}}\toprule
C3 trajectory & Static type & Round-0 type & Unchanged\\\midrule
Historical main & 4.99 & 3.83 & 7.83\\
Revised protocol & 5.08 & 3.87 & 7.77\\\bottomrule
\end{tabular}
\caption{Mean absolute target-failure trajectory error (pp). The historical and revised experiments are related and are not two independent validations.}
\end{table}

The static model predicts final target failures of 0.34\% versus 16.0\% observed in M1/tray/V2, 0.41\% versus 5.6\% in M1/stack/V2, and zero versus 3.2\% in M1/tray/V1. Within-type response is not stable across visited distributions: M1/tray/V2's ratio is $.450$ in diagnosis, $.821$ at round 0, and $.970$ at round 4. M1/stack/V1 instead changes from $.713$ in diagnosis to $.072$ at round 0. Later on-policy ratios are explanatory measurements, not information available to a pre-training forecast. The failed approximation limits type-average modeling, not the exact selection identity.

\balance
\section{Permutation Results and Other Counterexamples}
\label{sec:supp_permutation}
\begin{table}[t]
\centering\footnotesize\setlength{\tabcolsep}{4pt}
\begin{tabular}{@{}lrrrrr@{}}\toprule
Cell & Seed 0 & 1 & 2 & 3 & 4\\\midrule
M1 / stack / V1 & 1/0 & 0/0 & 0/0 & 1/1 & 0/0 \\
M1 / stack / V2 & 6/0 & 7/1 & 10/6 & 3/2 & 2/1 \\
M1 / tray / V1 & 7/2 & 2/1 & 2/8 & 4/3 & 1/1 \\
M1 / tray / V2 & 16/7 & 18/13 & 10/6 & 13/6 & 23/18 \\
J1 / stack / V1 & 11/2 & 1/1 & 4/7 & 4/2 & 1/1 \\
J1 / stack / V2 & 8/1 & 7/1 & 8/6 & 2/3 & 3/1 \\
J1 / tray / V1 & 15/0 & 0/0 & 10/0 & 31/6 & 21/0 \\
J1 / tray / V2 & 4/2 & 4/4 & 2/3 & 4/3 & 3/2 \\

\bottomrule\end{tabular}
\caption{Final target failure counts for C3/C2* per seed; each denominator is 100. Full success and other-failure seed rows accompany the manuscript as CSV. This table exposes small counts hidden by cell means.}
\end{table}

\paragraph{What permutation does and does not hold fixed.}
Within each C2* round, the judge labels its own current pool. A random permutation within true-success and true-failure strata preserves that pool's TP and FP exactly, but breaks their assignment to individual states. In later rounds the pool may differ from C3's pool, so the two trajectories may have different acceptance totals, error rates, and update strengths. The final evaluation counts physical failures regardless of whether the judge would accept them. This is a causal intervention on the learning procedure, but not identification of an isolated state-association coefficient at matched trajectory-wide noise.

The weakest cells matter: M1/stack/V1 has a $0.2\pp$ interval touching zero, M1/tray/V1 a $0.2\pp$ estimate with interval $[-3.2,3.2]$, J1/stack/V1 $1.6\pp$ $[-1.4,5.4]$, and J1/tray/V2 $0.6\pp$ $[-0.2,1.4]$. They are not proven null effects. The historical direct C3--C2 contrast also has unresolved conditions: intervals cross zero in J1/tray/V2 and M1/stack/V1. Static separability therefore cannot substitute for measuring the actual downstream effect.

\paragraph{Historical neural-policy pilot.}
A separate one-seed ManiSkill diffusion-policy pilot used 300 scripted stacking demonstrations and a wider-field $128\times128$ V2 camera. Its early checkpoint succeeded on 368/500 rollouts; failure types included 83 beside, 38 never grasped, 7 edge, 3 hover, and 1 behind. It barely sampled the hover failure central to the current mechanism. One round of filtered retraining gave success $.975$ for truth, $.975$ for DINOv2, $.970$ for J1, $.960$ for a policy-feature probe, and $.935$ for random filtering on 200 new scenes, versus $.775$ for the early checkpoint on that evaluation. These are descriptive one-seed values from a different evaluation pool, not evidence of equivalence or confirmation of the multi-seed mechanism. The planned neural confirmation remains unvalidated.

\paragraph{Reproducibility and reporting.}
The accompanying reproducibility materials contain seed summaries, rollout-level height and decision tables, exact figure-source records and hashes, plotting code, bibliography, and a source-to-claim index. Full raw scientific records and documented failed checks are retained with the study archive.

\end{document}